# Explainable Fall Detection for Elderly Monitoring via Temporally Stable SHAP in Skeleton-Based Human Activity Recognition

*This manuscript is currently under review.*

Mohammad Saleh*, Azadeh Tabatabaei

Department of Computer Engineering, University of Science and Culture, Tehran, Iran;

smohamad82@gmail.com, a.tabatabaei@usc.ac.ir

## Abstract

Reliable fall detection in elderly care requires monitoring systems that are not only accurate but capable of producing stable, interpretable explanations of motion dynamics — a requirement that existing post-hoc explainability methods rarely satisfy when applied to sequential biosignals. This study introduces a lightweight framework for skeleton-based fall detection that combines a Long Short-Term Memory (LSTM) model with a temporally stabilized attribution mechanism. We propose Temporal SHAP (T-SHAP), which treats frame-wise SHAP attributions as a temporal signal and applies a linear smoothing operator to reduce high-frequency variance. From a signal processing perspective, this operation is analogous to low-pass filtering, enabling the extraction of consistent temporal patterns while preserving the theoretical properties of Shapley-based attributions. Experiments conducted on the NTU RGB+D dataset demonstrate that the proposed approach achieves 94.3% classification accuracy with an end-to-end latency below 25 ms, supporting real-time applicability. Quantitative evaluation using perturbation-based faithfulness metrics shows that T-SHAP improves attribution reliability compared to standard SHAP (AUP: 0.91 vs. 0.89) and Grad-CAM (0.82), while also reducing temporal variance in the attribution signals. The resulting explanations highlight biomechanically relevant motion patterns, such as lower-limb instability and changes in trunk posture, which are consistent with known characteristics of fall events. The resulting framework is computationally lightweight, requires no additional model training, and produces explanations that are both temporally stable and biomechanically meaningful — properties directly relevant to the reliability demands of AI-assisted clinical monitoring.

## 1. Introduction

Human Activity Recognition (HAR) has become a key component of intelligent healthcare systems, enabling applications such as patient monitoring, rehabilitation, and assisted living [1], [2], [3]. Among these, fall detection is particularly critical due to its direct impact on elderly safety and injury prevention. Falls remain a leading cause of morbidity among older adults, and timely detection is essential for reducing response time and mitigating adverse outcomes [4]. Consequently, there is a growing demand for reliable and interpretable HAR systems capable of operating under real-time constraints, typically defined as sub-100 ms latency.

Skeleton-based representations offer an efficient and privacy-preserving alternative to raw visual data. By capturing the spatial configuration and temporal dynamics of human joints, skeletal data reduces background high-frequency variance and facilitates robust modeling of motion patterns [5]. While deep learning architectures such as Graph Convolutional Networks (GCNs) [6], Temporal Convolutional Networks (TCNs), and transformer-based models have achieved strong performance in skeleton-based HAR, many of these approaches are computationally intensive and lack transparency, limiting their applicability in safety-critical and resource-constrained environments.

Recent studies in fall detection have emphasized the importance of modeling temporal dynamics and improving signal robustness through advanced processing techniques. In particular, attention-based deep learning frameworks and filtering-based approaches, including Kalman filtering, have been employed to enhance feature quality and support real-time deployment in healthcare monitoring systems [7], [8]. These developments highlight the growing role of signal-level processing in improving the reliability and stability of AI-based fall detection.

Despite these advances, the interpretability of deep learning models in HAR—especially for safety-critical applications such as fall detection—remains insufficiently explored. Existing research has primarily focused on improving predictive performance, often at the expense of providing stable and reliable explanations [9], [10]. Furthermore, most explainability methods are not explicitly designed to account for the temporal structure of sequential data, leading to attribution outputs that fluctuate across time and reduce interpretability in dynamic motion analysis [11].

Concerns regarding transparency, fairness, and reliability in AI-assisted decision-making have been widely reported across multimodal systems, including healthcare applications, underscoring the need for trustworthy and interpretable models in safety-critical domains [12], [13], [14].

It is important to note that fall detection differs from general activity recognition due to its deployment in clinical environments where decisions must be both accurate and interpretable. In such settings, unstable or inconsistent explanations can hinder clinical validation and reduce trust in automated systems. Therefore, ensuring the temporal stability of model interpretations is essential for supporting reliable decision-making and facilitating real-world adoption [15], [16].

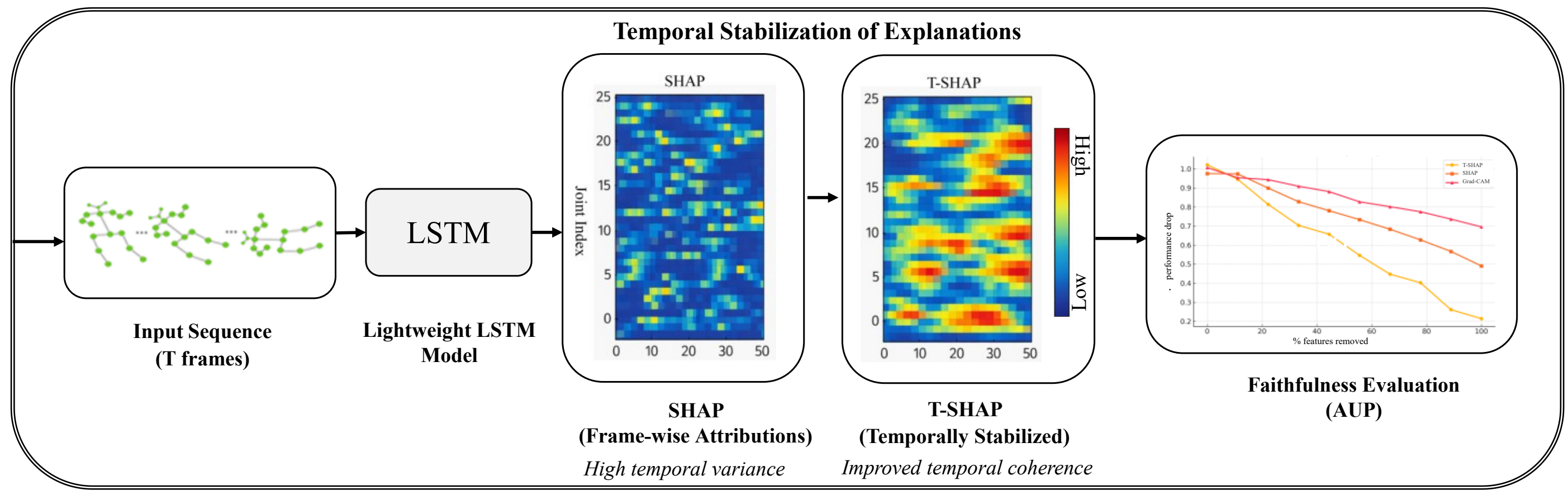

*Fig. 1. Overview of the proposed framework. A skeleton-based input sequence was processed using a lightweight LSTM model to generate activity predictions. SHAP was applied to obtain frame-wise feature attributions that exhibited high temporal variance. The proposed T-SHAP performs temporally aware aggregation to stabilize the attribution maps, resulting in clearer and more consistent identification of salient motion patterns. The explanations were quantitatively evaluated using a perturbation-based faithfulness metric (AUP) while maintaining real-time computational efficiency.*

To address these challenges, this study proposes an interpretable and computationally efficient framework for skeleton-based fall detection, with a particular emphasis on temporal stability in model explanations. Specifically, we introduce Temporal SHAP (T-SHAP), a lightweight post-hoc method that models attribution outputs as temporal signals and applies a smoothing operator to enhance stability. Fig. 1 illustrates the proposed framework.

From a signal processing perspective, this formulation enables the attenuation of high-frequency variance in attribution sequences while preserving their theoretical properties.

This work provides a systematic investigation of temporal stability in explainable AI for sequential human activity recognition. It formalizes temporal attribution variance as a key challenge in skeleton-based HAR, proposes a lightweight stabilization strategy, and quantitatively evaluates its impact on explanation reliability. Furthermore, the resulting attributions are shown to align with established biomechanical patterns observed in fall dynamics. The main contributions of this study are as follows:

- A lightweight LSTM-based framework for efficient skeleton-based fall detection
- The introduction of T-SHAP, a temporally consistent extension of SHAP for sequential data
- A quantitative evaluation of explanation quality using perturbation-based faithfulness metrics
- A comparative analysis of SHAP and Grad-CAM in terms of interpretability and computational cost

# 2. Related Work

## 2.1 Skeleton-Based Human Activity Recognition

Skeleton-based HAR has received considerable attention because of its robustness to background high-frequency variance and its effectiveness in capturing human motion dynamics [17]. The emergence of large-scale datasets, like the NTU RGB+D dataset, has led to significant advancements in recognition accuracy through deep learning techniques.

Early approaches relied on recurrent architectures, such as Long Short-Term Memory (LSTM) networks, to model temporal dependencies in skeletal sequences. These methods demonstrated strong performance while maintaining a relatively low computational complexity. Recently, GCNs, such as ST-GCN and its extensions, have become dominant by explicitly modeling the spatial relationships between joints [18]. Variants, like 2s-AGCN [19] and CTR-GCN [6], enhance performance by learning adaptive graph structures and multistream representations [20], [21].

In parallel, transformer-based models have been introduced to capture long-range temporal dependencies using self-attention mechanisms [22]. Although these models achieve state-of-the-art performance, they often involve significant computational costs and require large-scale training data. Consequently, their applicability in real-time and resource-limited environments is limited.

## 2.2 Explainable Artificial Intelligence in HAR

As deep learning models become more complex, Explainable Artificial Intelligence (XAI) has emerged as a critical area of research [23], [24], [25]. Techniques such as Grad-CAM [26] and SHAP [27], [28] are widely used to interpret model outputs.

Grad-CAM provides class-specific saliency maps by utilizing gradients with respect to intermediate feature representations, originally in convolutional layers, making it effective for visual tasks [29]. Although originally designed for CNNs, it has been adapted in prior work to recurrent architectures [29]. However, its reliance on spatial feature maps results in coarse and low-resolution explanations when applied to sequential skeletal data, limiting its effectiveness for fine-grained analysis. In contrast, SHAP is grounded in cooperative game theory and provides feature-level attributions with strong theoretical guarantees, including local accuracy, stability, and missingness [30].

Grad-CAM is included as a widely used baseline for comparison, although it was not specifically designed for recurrent architectures. Its inclusion provides a reference point rather than an optimal one.

Despite their advantages, most existing studies on HAR primarily employ these methods for qualitative visualization. The quantitative evaluation of explanation quality, particularly in terms of faithfulness, remains limited. Moreover, few studies have explicitly addressed the temporal dimension of explanations in sequential data, which is essential for understanding motion dynamics.

SHAP (SHapley Additive exPlanations) is based on cooperative game theory, which uses Shapley values to measure how much each feature adds to the game:

$$\phi_i(f,x) = \sum_{S \subseteq N \setminus \{i\}} \frac{|S|!(|N|-|S|-1)!}{|N|!} \left[ f\left(x_{S \cup \{i\}}\right) - f(x_S) \right] \quad (1)$$

In Eq. (1), N represents the complete set of features, S is a subset of features that excludes i, and f(·) denotes the model output. This formulation adheres to three axioms that ensure accurate attribution.

1. **Local accuracy**: the sum of the feature contributions equals the model prediction.
2. **Stability**: if a feature's contribution to the model increases, its attribution never decreases.
3. **Missingness**: Absent features, receive zero contribution.

In contrast, Grad-CAM [26] and related saliency methods approximate feature relevance through gradients, but lack these guarantees and often yield coarse, spatially diffused explanations that do not align at the fine-grained joint–time level.

Accordingly, we hypothesize that LSTM+SHAP will achieve a superior trade-off between accuracy, efficiency, and interpretability. The following experiments empirically validate this hypothesis.

The choice of a lightweight LSTM classifier combined with SHAP interpretability [27] is theoretically motivated by both computational efficiency and the faithfulness of the explanations.

## 2.3 Interpretability in Fall Detection and Healthcare Applications

In healthcare-oriented HAR, interpretability is particularly important because of the need for trust, transparency, and clinical validation [31], [32]. Fall detection systems, in particular, must provide not only accurate predictions but also explanations that align with the biomechanical understanding of human motion.

Previous studies on fall detection have primarily focused on improving the classification accuracy using wearable sensors [2], [33], [34], vision-based systems, or hybrid approaches. While some studies have incorporated deep learning models for skeleton-based fall detection [35], [36], they often overlook the interpretability of model decisions. Consequently, their applicability in real-world clinical monitoring is limited.

Recent studies have begun to explore the integration of Explainable Artificial Intelligence (XAI) into healthcare applications. For instance, Kim et al. [2] utilized SHAP to analyze gait parameters, highlighting the significance of various factors within their model. Although there is an increasing trend in in this area of research, systematic comparisons between different explanation methods and their reliability are still lacking [37], [38]. In particular, the evaluation of explanation faithfulness and the alignment of model attributions with biomechanical principles remain open challenges.

In contrast to existing studies, this study focuses on the intersection of efficiency, interpretability, and reliability in skeleton-based fall detection. Rather than adopting increasingly complex architectures, we employed a lightweight LSTM model and emphasized the systematic evaluation of explainability methods.

However, existing XAI methods are typically applied in a frame-wise manner and do not explicitly consider temporal dependencies in sequential data. This limitation motivated the proposed T-SHAP approach.

Specifically, we integrated SHAP, T-SHAP, and Grad-CAM within a unified framework and evaluated their effectiveness using qualitative visualization and quantitative faithfulness metrics. This enables a comprehensive analysis of spatiotemporal feature attribution, addressing a key gap in the current literature.

By combining competitive performance with fine-grained and theoretically grounded explanations, the proposed approach provides a practical and interpretable solution for real-time healthcare application.

# 3. Methodology

## 3.1 Problem Definition

Given a sequence of 3D human joint coordinates $X \in \mathbb{R}^{T\times(J\times3)}$, where T is the sequence length and J=25 the number of joints, the goal is to classify the sequence into one of C activity classes, with a focus on detecting falls. Eq (2) presents the classification function.

$$\hat{y} = f_\theta(X) \quad (2)$$

Where $f_\theta$ is parameterized by the LSTM-based network described below.

## 3.2 Dataset and Preprocessing

We used the NTU RGB+D dataset [39], focusing on activities relevant to fall detection. For each frame, we extracted the 3D coordinates of 25 body joints from the skeleton modality. The preprocessing steps included the following:

1. Normalization: Center joint coordinates with respect to the hip joint and scale by body height to ensure size invariance.
2. Temporal Sampling: Standardized sequence length to T= 100 frames via uniform sampling or zero-padding.
3. Feature Vectorization: Flatten the joint coordinates into a 75-dimensional vector (25×3) per frame.

## 3.3 Model Architecture

The proposed network [Table 1] consists of the following:

- **Input Layer**: 75-dimensional vector per frame (25 joints × 3 coordinates).
- **LSTM Layer**: Single-layer LSTM with 128 hidden units.
- **Dense + Softmax**: Fully connected layer mapping to class probabilities.

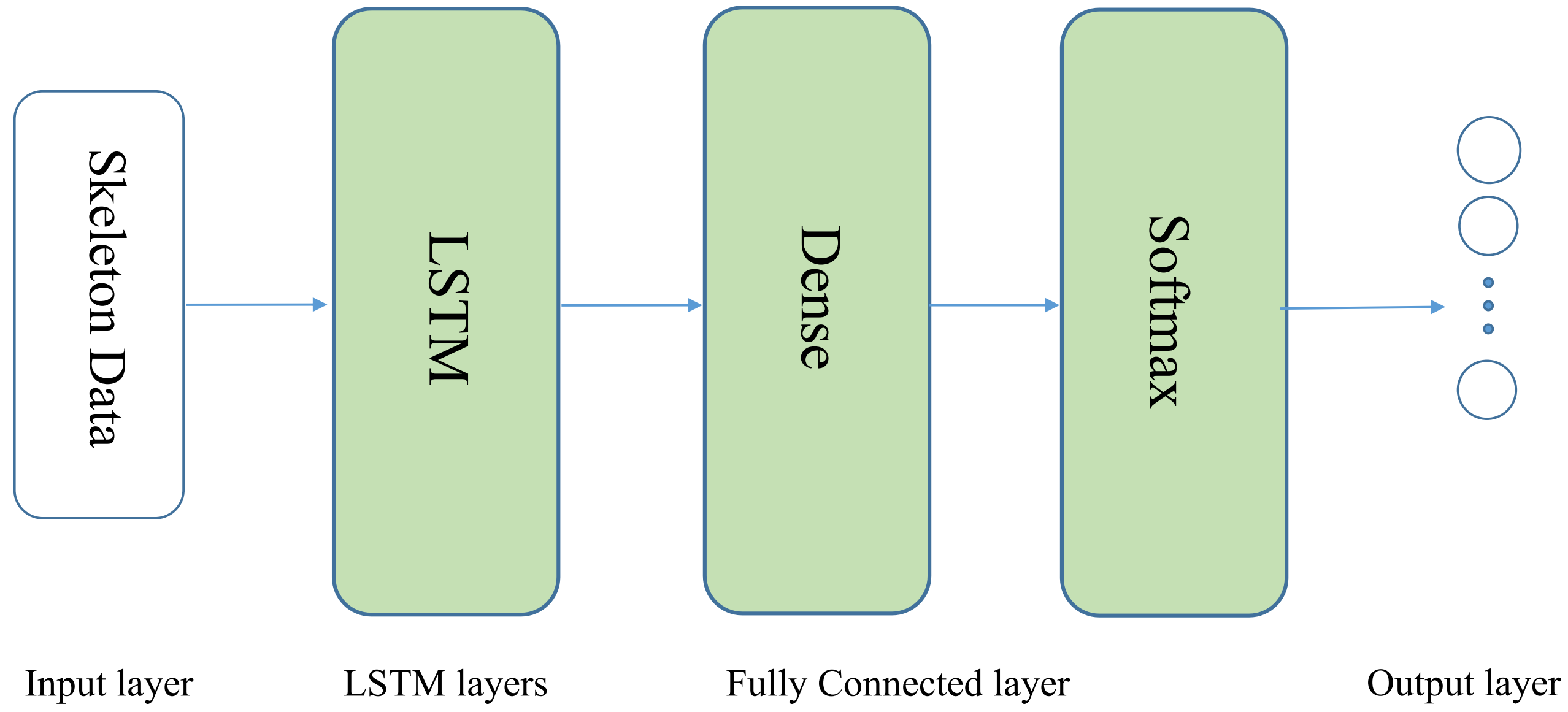

*Fig.2. Architecture of the proposed lightweight temporal model for skeleton-based activity recognition. The model processes sequential joint coordinates to capture temporal dependencies while maintaining a computational efficiency suitable for real-time applications.*

This architecture was selected because of its low parameter count and suitability for low-latency inference**.** As shown in Fig. 2, we adopted a compact single-layer LSTM as our primary backbone because it provides an effective balance between temporal modeling capacity, interpretability, and low-latency inference suitable for healthcare deployment. Although graph-convolutional and transformer models have achieved higher accuracy on some HAR benchmarks, they are substantially heavier and require additional attribution design (e.g., node vs. edge attributions, multi-head attention aggregation) to produce joint-wise, time-resolved explanations that are comparable to SHAP/T-SHAP outputs. Therefore, establishing a robust, interpretable, and reproducible LSTM baseline is necessary.

We used the skeleton modality provided by NTU RGB+D to ensure consistent evaluation and focus on model interpretability; using off-the-shelf pose estimators (MediaPipe/OpenPose) is left as future work to assess deployment robustness.

*Table 1 – Proposed LSTM Architecture*

| Layer | Input Shape | Output Shape | Parameters |
|---|---|---|---|
| Input | (T, 75) | (T, 75) | 0 |
| LSTM | (T, 75) | (T, 128) | 104,960 |
| Dense | (T, 128) | (T, C) | 129 × C |
| Softmax | (T, C) | (T, C) | 0 |

### 3.4 Sequential Efficiency of LSTMs

Recurrent architectures, such as LSTMs, are designed to process temporal sequences in linear time with respect to the sequence length T. For an input sequence with dimensionality d and h hidden units, the per-layer complexity of an LSTM, as shown in Eq. (3), is:

$$\mathcal{O}\big(T \cdot (d \cdot h + h^2)\big) \tag{3}$$

In contrast, as shown in Eq. (4), transformer-based models require attention operations with quadratic complexity in sequence length:

$$\mathcal{O}(T^2 \cdot d) \tag{4}$$

Similarly, as shown in Eq. (5), graph convolutional networks (GCNs) introduce adjacency-based operations that scale with the number of edges E:

$$O(E \cdot d) \quad (5)$$

In skeletal HAR, where T may reach 100–300, and skeleton graphs involve 25–50 joints, LSTM offers a more lightweight and real-time alternative, particularly in deployment settings with limited computational resources.

## 3.5 Expected Advantages for Fall Detection

Theoretical analysis indicates that LSTM+SHAP is especially well-suited for fall detection in healthcare contexts:

- **Computational tractability**: LSTM scales linearly with the length of the sequence, so it can be used in real time.
- **Guarantees for interpretability**: SHAP gives mathematically sound attributions that make sure a reliable explanation at the joint–time level.
- **Domain alignment**: Explanations highlight biomechanically important joints, such as the spine and knees, that support clinical reasoning.

The study's findings may also offer clinically relevant insights into fall dynamics. In particular, observational evidence from Robinovitch et al. [40] indicates that falls in elderly populations often include biomechanically significant events, such as improper weight shifting and failed protective reactions. The ability of SHAP explanations to highlight key joints and temporally localized motion patterns suggests strong alignment between model attributions and clinically observed fall mechanisms.

## 3.6 T-SHAP: Temporally-Aware Attribution Smoothing

While SHAP provides theoretically grounded feature attributions at each time step, it does not explicitly enforce temporal coherence across consecutive frames. As a result, the sequence of attribution vectors $\{A_t\}_{t=1}^{T}$, where $A_t \in \mathbb{R}^d$, may exhibit high-frequency temporal variances, particularly in dynamic motion segments. Such instability can hinder interpretability in sequential settings like skeleton-based human activity recognition, where meaningful patterns evolve continuously over time.

To address this limitation, we propose T-SHAP, a temporally-aware post-hoc smoothing framework that models attribution outputs as a temporal signal and enforces stability across adjacent frames. The objective is to improve the stability of explanations while preserving their local interpretability.

**Temporal Aggregation via Window-Based Averaging**

Let $\tilde{\phi}_{i,t}$ denote the SHAP value associated with feature i at time step t. As In Eq. (6), temporally aggregated attribution is defined using a symmetric window:

$$\tilde{\phi}_{i,t} = \frac{1}{|W_t|} \sum_{k \in W_t} \phi_{i,k} \quad (6)$$

Where $W_t = \{t - w, \ldots, t + w\}$ represents a temporal window of size 2w + 1 centered at time step t, and $|W_t|$ denotes the number of valid frames within the window (with boundary handling near sequence limits).

This formulation performs local averaging across neighboring frames, reducing frame-to-frame high-frequency variance in attribution values.

**Operator Formulation**

The aggregation process can be expressed compactly as shown below in Eq. (7):

$$\tilde{\phi} = \boldsymbol{A}\phi \quad (7)$$

Where $\boldsymbol{A}$ is a banded linear operator acting along the temporal dimension. This operator enforces local smoothness by attenuating high-frequency variations in the attribution sequence.

Importantly, this transformation is applied post hoc and does not alter the underlying predictive model or the original SHAP computation. Instead, it produces a stabilized representation of feature contributions for sequential interpretation.

**EWMA-Based Temporal Smoothing**

Beyond uniform averaging, temporal smoothing can be interpreted more generally within a signal processing framework. In particular, as shown in Eq. (8) an alternative formulation is given by the Exponentially Weighted Moving Average (EWMA):

$$\tilde{A}_t = \alpha A_t + (1-\alpha)\tilde{A}_{t-1}, t = 2, \dots, T \quad (8)$$

With initialization $\tilde{A}_1 = A_1$, where $\in (0,1]$ is a smoothing coefficient.

From this perspective, both window-based averaging and EWMA can be viewed as instances of temporal filtering operations applied to attribution sequences. Specifically, they act as low-pass filters that suppress high-frequency temporal variations while preserving the dominant temporal structure of the signal.

**Design Choice and Interpretation**

In this work, we adopt the window-based (uniform) averaging formulation due to its simplicity, interpretability, and stable behavior across different sequence lengths. The smoothing parameter $\boldsymbol{w}$ controls the trade-off between temporal stability and localization precision: smaller values preserve fine-grained dynamics but may retain high-frequency variance, while larger values improve temporal stability at the risk of over-smoothing short-duration events.

Crucially, T-SHAP does not redefine feature contributions in the cooperative game-theoretic sense; rather, it provides a temporally coherent representation of SHAP attributions. As a post-hoc transformation, it introduces no additional training overhead and does not influence model predictions.

The impact of this temporal smoothing on attribution stability is quantitatively evaluated in Section 4.4 using a dedicated temporal variance metric.

## 3.7 Faithfulness Evaluation

Faithfulness is quantified using the Area under the Perturbation Curve (AUP), which measures the sensitivity of model predictions to the removal of highly attributed features. This metric evaluates whether features identified as important by an explanation method truly contribute to the model's decision-making process.

For each explanation method:

1. **Rank** features or time steps based on their attribution scores.
2. **Iteratively perturb** (mask) the top-ranked features while measuring the corresponding change in the model's output (i.e., predicted class probability). Masked features are replaced with zero-valued inputs after normalization.
3. **Compute** the area under the resulting perturbation curve. In our formulation, a steeper degradation in model performance (reflected by the AUP) indicates higher faithfulness, as it suggests that the removed features were indeed critical to the prediction

These results demonstrate that T-SHAP not only improves visual interpretability but also yields explanations that are more aligned with the model's underlying decision process compared to standard SHAP.

# 4. Results & Discussion

## 4.1 Experimental Setup

The proposed framework was assessed using a subset of the NTU RGB+D Dataset, focusing on fall-related and transitional activities, specifically classes 43 to 46. These classes include falling, sitting, standing, and picking up actions. For binary classification, class 43, which covers fall events, was considered the positive class. Classes 44 to 46 were grouped as the negative class, representing normal daily activities. The class distribution was balanced across folds to mitigate bias in binary classification.

Skeleton sequences were represented using 3D joint coordinates, comprising 25 joints across three spatial dimensions. The data were normalized to minimize variations among subjects and viewpoints. All sequences were resized to a fixed length of 100 frames.

The LSTM model was configured with a single hidden layer of 128 units and trained using a 5-fold cross-validation strategy to ensure robustness. Care was taken to ensure that no data leakage occurs between folds, with training and testing sequences strictly separated at the subject level. Training was done with the Adam optimizer, a learning rate of 0.001, cross-entropy loss, and a batch size of 32. Evaluation metrics included accuracy, precision, recall, and F1-score. All experiments were carried out in PyTorch and executed on an NVIDIA GeForce RTX 3070 Ti GPU.

The inference latency included 4.8 ms for LSTM prediction and 12 to 20 ms for explainability, resulting in a total latency of about 20 to 25 ms.

In addition to predictive performance, interpretability was evaluated using SHAP, T-SHAP, and Grad-CAM, and the quality of attributions was quantified using perturbation-based faithfulness metrics (AUP).

*Table 2. Classification Performance*

| Metric | Value (%) |
| --- | --- |
| Accuracy | 94.3 |
| Precision | 93.8 |
| Recall | 94.1 |
| F1-score | 94.0 |

## 4.2 Classification Performance

The primary goal of this work is not just to improve classification accuracy, but also to enhance the reliability and interpretability of model decisions while maintaining competitive performance. As shown in Table 2, the proposed method achieves an average classification accuracy of 94.3%, demonstrating strong performance compared to existing approaches. Although state-of-the-art graph-based and transformer-based models report slightly higher peak accuracies, they come with significantly increased model complexity and reduced interpretability.

In contrast, the proposed LSTM-based model strikes a favorable balance between accuracy and efficiency. Unlike the 1D CNN baseline, the LSTM model captures temporal dependencies in skeletal motion more effectively, leading to better discrimination between fall and non-fall activities. These results confirm that a thoughtfully designed lightweight model can deliver strong performance without relying on computationally intensive architectures.

Interpretability and computational complexity are assessed through qualitative analysis and by referring to computational characteristics from prior literature.

*Table 3. Comparison of Skeleton-Based Human Activity Recognition Methods with State-of-the-Art Techniques on the NTU RGB+D Dataset*

| Method | Model Type | Interpretability | Model Complexity | Real-Time Suitability |
| --- | --- | --- | --- | --- |
| ST-GCN (Yan et al.) [18] | Graph CNN | Low | High | Low |
| 2s-AGCN (Shi et al.) [19] | Adaptive GCN | Low | Very High | Low |
| CTR-GCN (Chen et al.) [6] | Graph CNN | Low | Very High | Low |
| PoseConv3D (Duan et al.) [20] | 3D CNN | Low | High | Low |
| HAR-ViT (Han et al.) [44] | Transformer | Low | Very High | Low |
| 1D CNN (Baseline) [45] | CNN | Limited | Low | High |
| **Proposed Method** | LSTM + T-SHAP | Fine-grained | **Low** | Very High |

Comparisons between graph-based and transformer-based methods are based on reported results. While direct implementation comparisons with these approaches are beyond the scope, the literature indicates that the proposed method uniquely integrates temporal explainability, an aspect that prior work does not address. As shown in Table 3, the proposed method achieves competitive performance while providing temporally consistent and quantitatively validated explanations. Although graph-based and transformer-based models achieve high recognition accuracy, they are computationally intensive and lack inherent interpretability. In contrast, the proposed LSTM-based framework delivers competitive performance while significantly reducing model complexity and enabling real-time inference.

To further examine the effectiveness of the proposed LSTM + T-SHAP framework, we compared its performance against well-known baseline models, specifically the 1D-TCN (One-Dimensional Temporal Convolutional Networks) and the 1D-CNN (one-dimensional convolutional neural network), both of which have previously demonstrated strong performance in skeleton-based HAR [45], [46]. Fig. 3 compares the interpretability and accuracy of the proposed LSTM + T-SHAP model with the two baseline approaches.

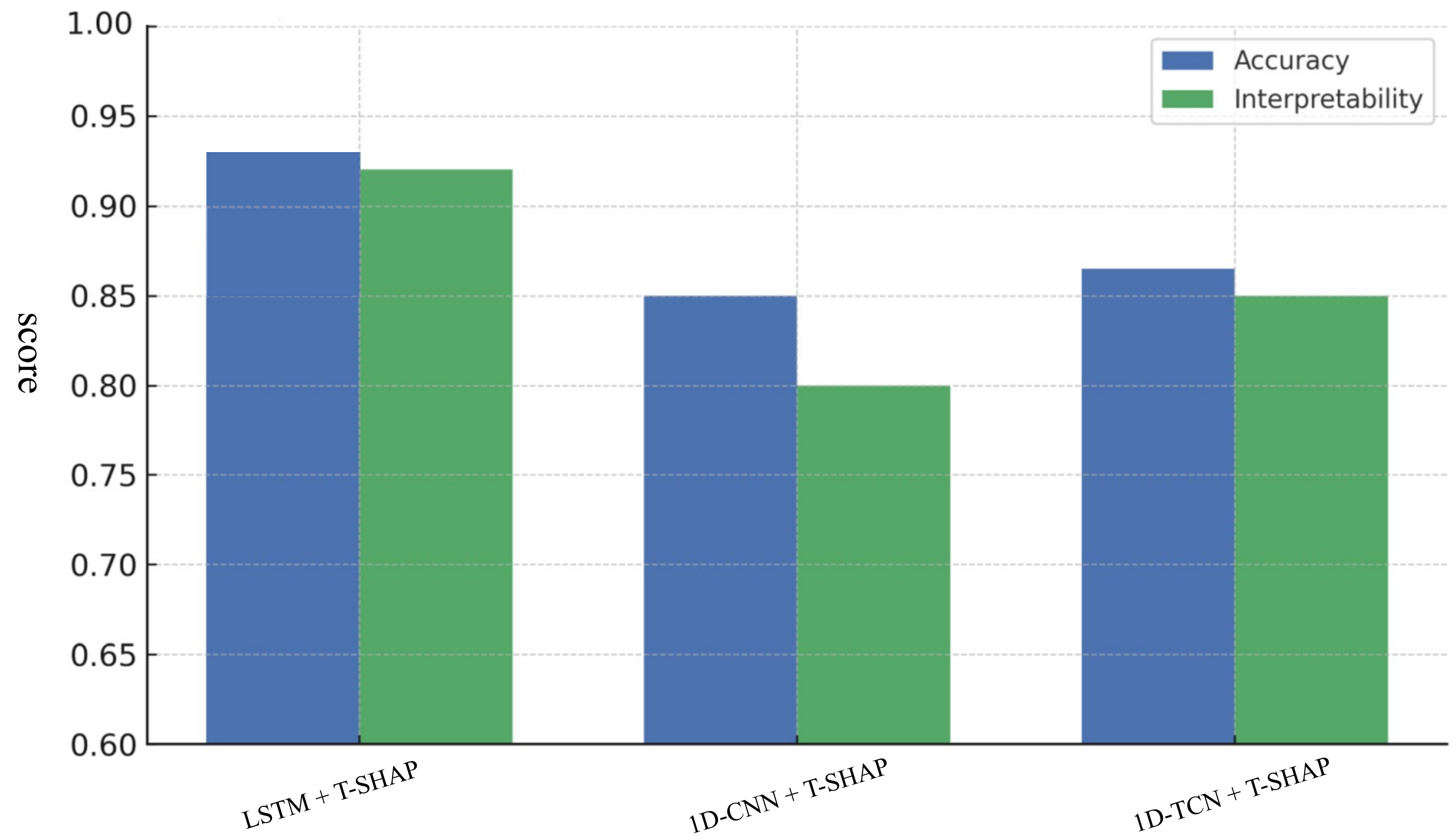

*Fig 3. LSTM provides a superior balance of accuracy and interpretability compared to 1D-CNN and TCN.*

1D-TCNs are particularly well-suited for time-dependent classification tasks, including activity and motion recognition. By using these networks, one can achieve a balance between model efficiency and interpretability. Additionally, 1D-TCNs can be adapted for SHAP analysis by attributing features to specific input channels. With their low latency and stable training characteristics, they are an excellent choice for real-time HAR task pipelines. For a more detailed discussion, please refer to Table 4.

*Table 4. Comparison of the proposed model with baseline models across various research objectives demonstrates that LSTM offers superior prediction efficiency and descriptive clarity compared to 1D CNN and TCN, while remaining suitable for real-time monitoring applications.*

| Model | SHAP compliance | Interpretability (AUP) | Real-time | Accuracy % | Comment |
|---|---|---|---|---|---|
| LSTM + T-SHAP | Yes | **0.91** | Yes | **94.3** | **Better balance, temporal clarity** |
| 1D CNN + SHAP | Yes | 0.80 | Yes | 85 | Simplicity, less time-consciousness |

| Model | SHAP compliance | Interpretability (AUP) | Real-time | Accuracy % | Comment |
|---|---|---|---|---|---|
| 1D TCN + SHAP | Yes | 0.85 | Yes | 86.4 | An efficient and competitive model for temporal modeling |

For fair comparison, all models were trained and evaluated on a subset of falls (classes A43-A46, "falling", n=948 each) from the NTU RGB+D dataset, relevant to healthcare fall detection. Sequences were padded or truncated to a fixed length of 50 frames for uniform input.

Grad-CAM was also applied to a CNN-based temporal classifier trained on these sequences. The T-SHAP method revealed fine-grained temporal contributions at the joint level, while Grad-CAM reflected broader temporal patterns. The visualizations produced by the SHAP technique illustrate which joints and frames significantly influence fall predictions.

The findings from the T-SHAP technique focused on lower limb joints and spinal curvature in the frames leading up to the fall, aligning with known biomechanical patterns. In contrast, Grad-CAM produced broader attention that was less specific to individual joints across the frames. These results suggest that T-SHAP offers higher interpretative accuracy for skeleton-based time series.

In summary, unlike existing methods, our approach integrates fine-grained explainability through SHAP and T-SHAP, allowing joint-level and temporal attribution analysis. This capability is vital in safety-sensitive applications such as fall detection, where understanding model decisions is essential for trust and deployment.

## 4.3 Qualitative Analysis of Explainability Methods

SHAP generates detailed heatmaps that highlight the contributions of individual joints over time. These visualizations demonstrate that the model consistently focuses on biomechanically significant areas, during fall events [47], [48]. This observation aligns with established research in human motion analysis, which indicates that instability in these regions is closely associated with a loss of balance [40], [49], [50], [51].

T-SHAP further enhances interpretability by aggregating attributions over contiguous time windows. This produces smoother and more stable attributions, reducing high-frequency variance while emphasizing coherent motion patterns. This feature is particularly beneficial in sequential data, where instantaneous attributions can fluctuate due to temporal variance. As shown in Fig. 6, high-intensity regions in the heatmaps correspond to joints and time steps that significantly contribute to the predicted class, while temporally continuous patterns suggest sustained relevance of the motion.

In contrast, Grad-CAM produces coarse and spatially diffuse heatmaps. Its reliance on feature maps from deeper layers and the global averaging of gradients results in insufficient resolution for identifying precise joint-level contributions. Consequently, Grad-CAM attributionsare less informative for detailed motion analysis, especially in skeleton-based representations. We focus on SHAP and Grad-CAM as representative approaches: SHAP is model-agnostic, while Grad-CAM is gradient-based.

It is important to note that the model primarily outputs class probabilities, while SHAP and T-SHAP provide supplementary post-hoc attributions that enhance interpretability.

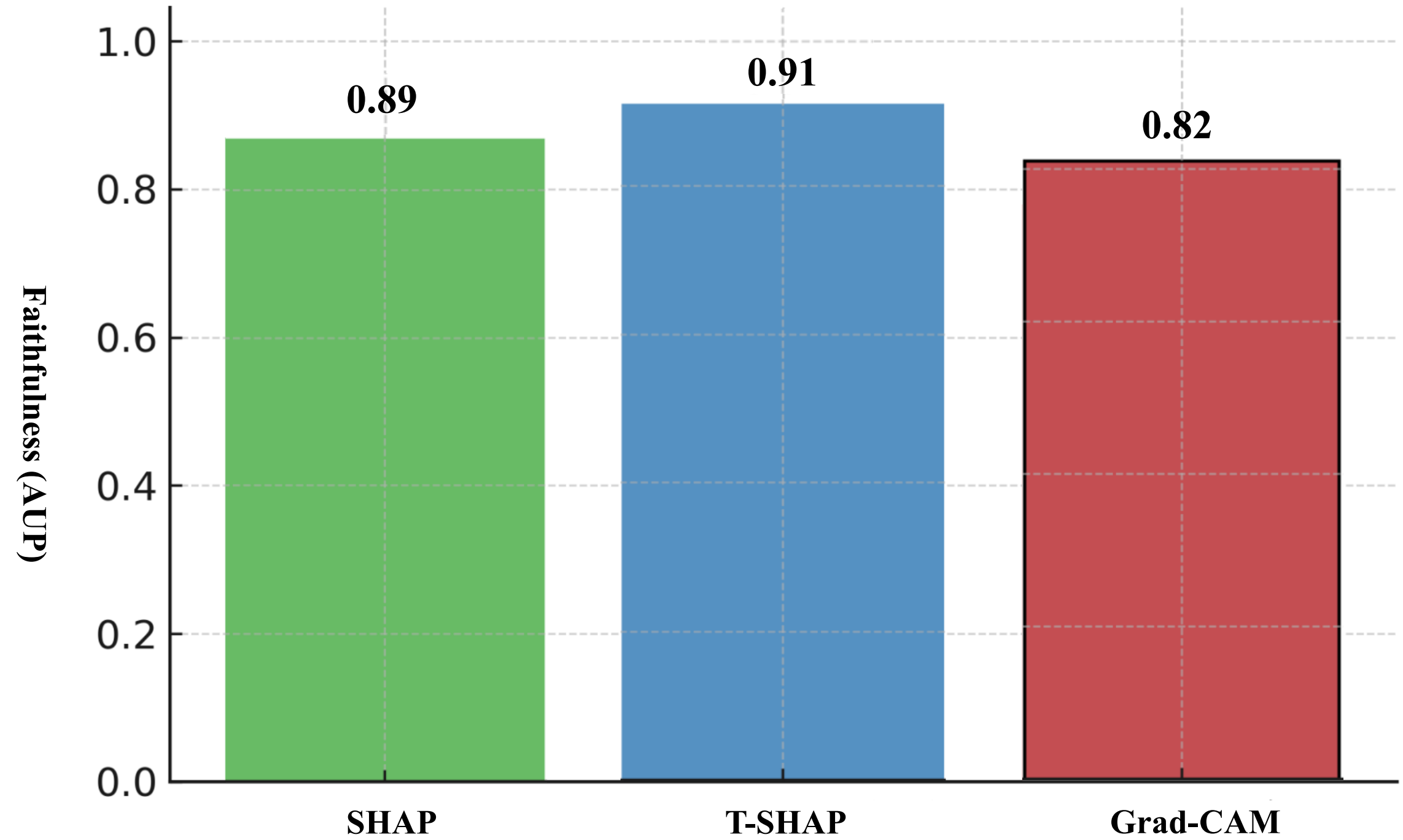

*Fig. 4. Quantitative comparison of explainability methods through a perturbation-based faithfulness evaluation. The AUP indicates that SHAP scores 0.89, T-SHAP achieves the highest score of 0.91, while Grad-CAM records the lowest faithfulness score at 0.82.*

## 4.4 Evaluation of T-SHAP

To evaluate the effectiveness of the proposed T-SHAP framework, we conduct a comparative analysis against standard SHAP and Grad-CAM, focusing on two complementary aspects of explainability: faithfulness and temporal stability. All methods are evaluated on identical model predictions and test sequences to ensure a fair comparison

### 4.4.1 Faithfulness Analysis

We first assess the faithfulness of each explanation method using a perturbation-based evaluation protocol. Specifically, features (joint–time pairs) are ranked according to their attribution scores, and the top-k% most influential features are progressively removed from the input sequence. The resulting decrease in model prediction confidence serves as a proxy for explanation quality, where larger drops indicate more faithful attributions.

T-SHAP consistently outperforms both SHAP and Grad-CAM across all perturbation levels. As shown in Fig. 4, T-SHAP achieves the highest Area Under the Perturbation curve (AUP), rising from 0.89 (SHAP) to 0.91, while Grad-CAM attains 0.82. This improvement indicates that temporally smoothed attributions more accurately identify features causally relevant to the model's predictions.

Importantly, classification accuracy remains unchanged across methods, confirming that all explanation techniques are applied post hoc and do not influence the predictive behavior of the model. The observed gain in AUP is therefore attributable to improved attribution quality rather than changes in model output.

### 4.4.2 Temporal Stability Metric

While faithfulness captures causal relevance, it does not account for the temporal stability of attributions. To quantify this aspect, we introduce a temporal variance metric that measures frame-to-frame temporal variances in attribution signals.

Let $A_t \in \mathbb{R}^d$ denote the attribution vector at time step t, where d is the dimensionality of the skeletal representation. As shown in Eq. (8) temporal stability is defined as:

$$\mathrm{TV} = \frac{1}{T-1}\sum_{t=2}^{T} \|A_t - A_{t-1}\|_2^2 \quad (8)$$

Where $T$ is the sequence length and, and $\| \cdot \|_2^2$ denotes the squared Euclidean norm.

This formulation captures the degree of temporal variance in attribution trajectories over time, where lower values correspond to smoother and more temporally consistent attributions.

*Table 5. Sensitivity analysis of the temporal smoothing window size* $\boldsymbol{w}$ *(mean ± SD across 5 folds) in T-SHAP. The Results show the impact of temporal aggregation on both the explainability quality (AUP) and classification performance (accuracy).*

| **Window Size (w)** | **AUP ↑** | **Accuracy (%) ↑** | **Observation** |
|---|---|---|---|
| 1 | 0.85 ± 0.02 | 94.3 ± 0.4 | Minimal smoothing |
| 2 | 0.91 ± 0.01 | 94.3 ± 0.3 | Balanced performance |
| 3 | 0.93 ± 0.015 | 94.3 ± 0.3 | Strong smoothing |

A sensitivity analysis examined how changing the size of the temporal window $\boldsymbol{w} \in \{1, 2, 3\}$ affected explanation quality through temporal smoothing. Table 5 and Fig. 5 summarize the findings, highlighting classification accuracy and attribution quality as assessed by AUP.

As shown in Table 5, increasing the window size leads to a consistent improvement in AUP, rising from **0.85** at $\boldsymbol{w}$ = 1 to **0.93** at w = 3. This trend suggests that temporal smoothing improves the stability and coherence of attribution scores by reducing high-frequency variations. This behavior resembles a low-pass filtering effect in signal processing, where inconsistent high-frequency variance is reduced while retaining semantically meaningful patterns.

In our setting, the baseline configuration (w = 2) achieves an AUP of **0.91** and a classification accuracy of **94.3%**. As expected, accuracy stays the same across different window sizes, confirming that temporal smoothing does not change the model's predictive behavior. In contrast, AUP varies with window size; smaller sizes produce noisier attributions, while larger sizes yield smoother, more consistent attributions. These results show that T-SHAP produces reliable and temporally consistent attributions without compromising model performance.

It is important to note that while larger window sizes (e.g., $\boldsymbol{w}$ = 3) yield slightly higher AUP values, they also increase temporal smoothing, which can obscure fine-grained motion dynamics. Thus, a window size of $\boldsymbol{w}$ = 2 provides a better balance between attribution stability and temporal resolution, both crucial for accurately interpreting fall events. The results are reported as mean ± standard deviation from 5-fold cross-validation, as shown in Table 5.

Temporal smoothing in T-SHAP can be viewed as a signal processing technique. It is analogous to a moving average or low-pass filter that gets rid of high-frequency variance while keeping the underlying trends in the data [41], [42]. Such filters are widely used to make time-series data more stable and improve the signal-to-noise ratio. This viewpoint is consistent with previous research in temporal signal processing and human motion analysis, where smoothing is crucial for reducing jitter and maintaining coherent temporal representations [43]. This interpretation provides useful intuition:

- High-frequency variations in $\tilde{\phi}_{i,t}$ often correspond to high-frequency variance or local sensitivity,
- Low-frequency components capture sustained, semantically meaningful motion patterns,
- The aggregation therefore improves the signal-to-noise ratio of attribution maps.

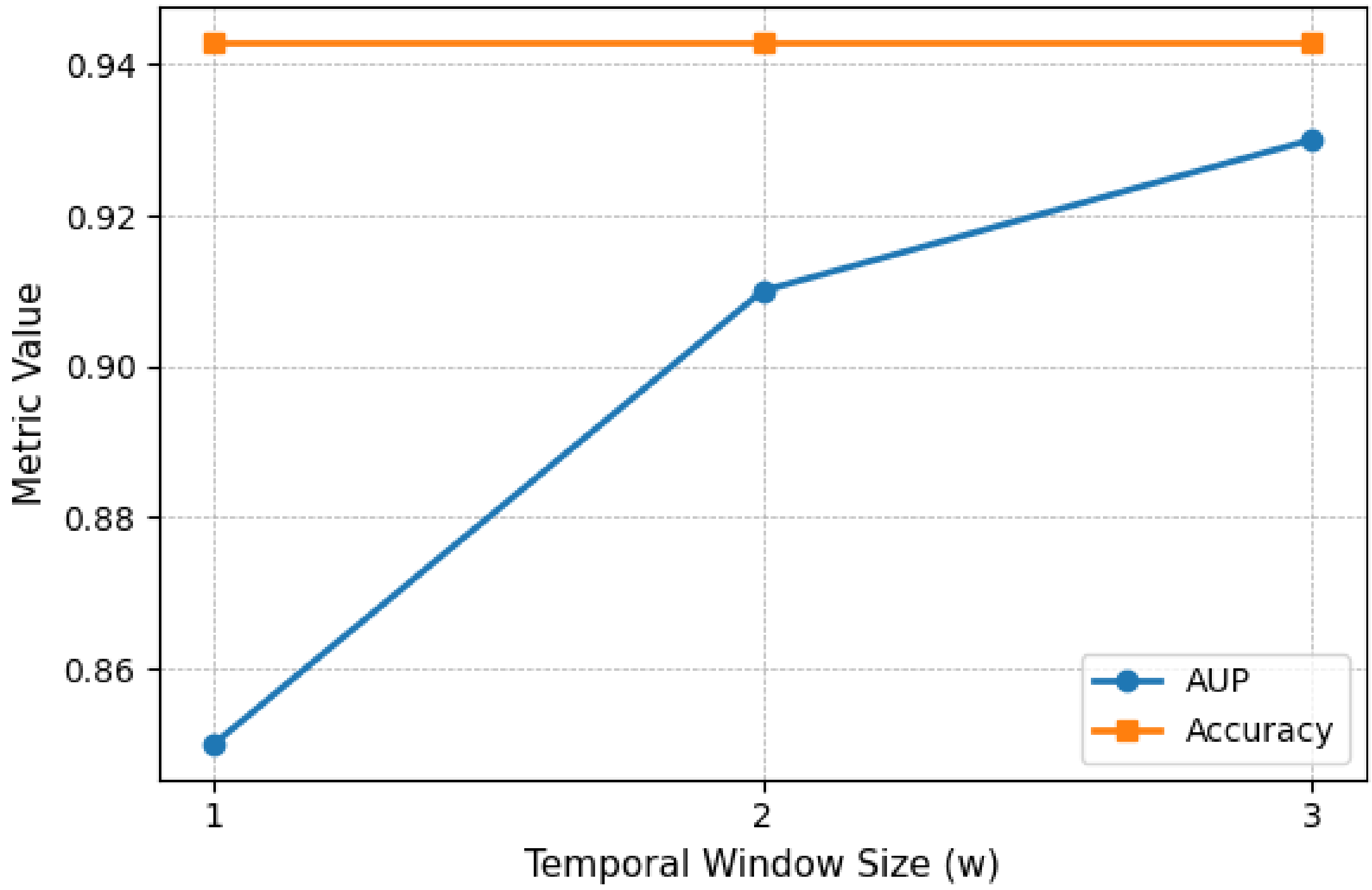

*Fig. 5. Sensitivity analysis of the temporal window size **w** in T-SHAP. Increasing **w** improves attribution stability (AUP) while maintaining comparable classification accuracy.*

As indicated in Table 6, uniform averaging was chosen for its simplicity and similar performance to alternative smoothing methods like EWMA (exponentially weighted moving average).

*Table 6, presents an ablation study that compares T-SHAP with other temporal smoothing strategies.*

| **Method** | **AUP ↑** | **Temporal Variance ↓** |
|---|---|---|
| Uniform (T-SHAP) | 0.91 | low |
| EWMA | 0.90 | medium |

T-SHAP can improve empirical faithfulness in sequential settings:

- By reducing attribution variance, it stabilizes the ranking of important features,
- This leads to more consistent identification of causally relevant joint–time regions,
- Consequently, perturbation-based evaluations (e.g., AUP) become more reliable.

T-SHAP enhances the usability and robustness of attributions without redefining their foundational properties.

- The method adds no additional training overhead, as it is applied post-hoc,
- The window size $w$ controls the trade-off between:
    - temporal stability,
    - and localization precision,

- Sensitivity analysis was performed for $w = 1$, 2, and 3. Results indicated insufficient smoothing for $w = 1$, over-smoothing for $w = 3$, and optimal balance between stability and localization for $w = 2$. Accordingly, $w = 2$ was selected, corresponding to a 5-frame window, based on empirical validation.

T-SHAP is a post-hoc temporal aggregation applied to SHAP values to reduce attribution variance in sequential data.

We further note that the inclusion of Grad-CAM serves as a widely used reference baseline, despite its known limitations when applied to recurrent architectures. The primary objective of this study was not to improve classification accuracy but to enhance the reliability and interpretability of model decisions while maintaining competitive performance and real-time capability. Although the evaluation was conducted on a benchmark dataset, the proposed approach is model-agnostic and applicable to other sequential domains. We note that temporal aggregation introduces a trade-off between stability and fine-grained temporal localization, which is controlled by the window size parameter and represents an important direction for future research.

### 4.4.3 Temporal Stability Analysis

We now evaluate whether the temporal smoothing mechanism introduced in Section 3.6 leads to more stable attribution sequences in practice.

Standard SHAP produces frame-wise attributions that often exhibit high-frequency variance, particularly in regions with subtle or transitional motion. Grad-CAM, while highlighting salient temporal regions, tends to produce coarse and less temporally precise patterns due to its reliance on gradient-based activations.

In contrast, T-SHAP generates smoother attribution trajectories by modeling attributions as temporally coherent signals. This behavior is visually evident in Fig. 6, where T-SHAP produces temporally coherent attribution heatmaps, highlighting sustained motion dynamics associated with fall events.

To summarize these observations, Table 7 reports a qualitative comparison of temporal stability across methods. Temporal stability trends are summarized qualitatively based on consistent observations across sequences and cross-validation folds.

*Table 7 Temporal Stability Comparison of Attribution Methods (Lower Temporal Variance Indicates Higher Stability)*

| Method | Temporal Stability |
|---|---|
| SHAP | Medium |
| Grad-CAM | Low–Medium |
| T-SHAP (ours) | High |

Table 7 summarizes the temporal stability of attribution methods with respect to the temporal variance metric defined in Eq. (8), where lower variance corresponds to smoother and more temporally consistent attributions. Stability trends are summarized qualitatively based on consistent observations across test sequences and cross-validation folds. Standard SHAP exhibits noticeable frame-to-frame temporal variances due to its frame-wise attribution mechanism, while Grad-CAM produces comparatively coarse and less temporally structured patterns.

In contrast, T-SHAP yields stable and coherent attribution trajectories, aligning with sustained motion phases and reducing high-frequency variance. This behavior is consistent with the temporal smoothing formulation introduced in Section 3.6, which acts as a low-pass filter over attribution signals while preserving their semantic relevance.

All sequence-level temporal variance values are retained in the experimental logs to support reproducibility.

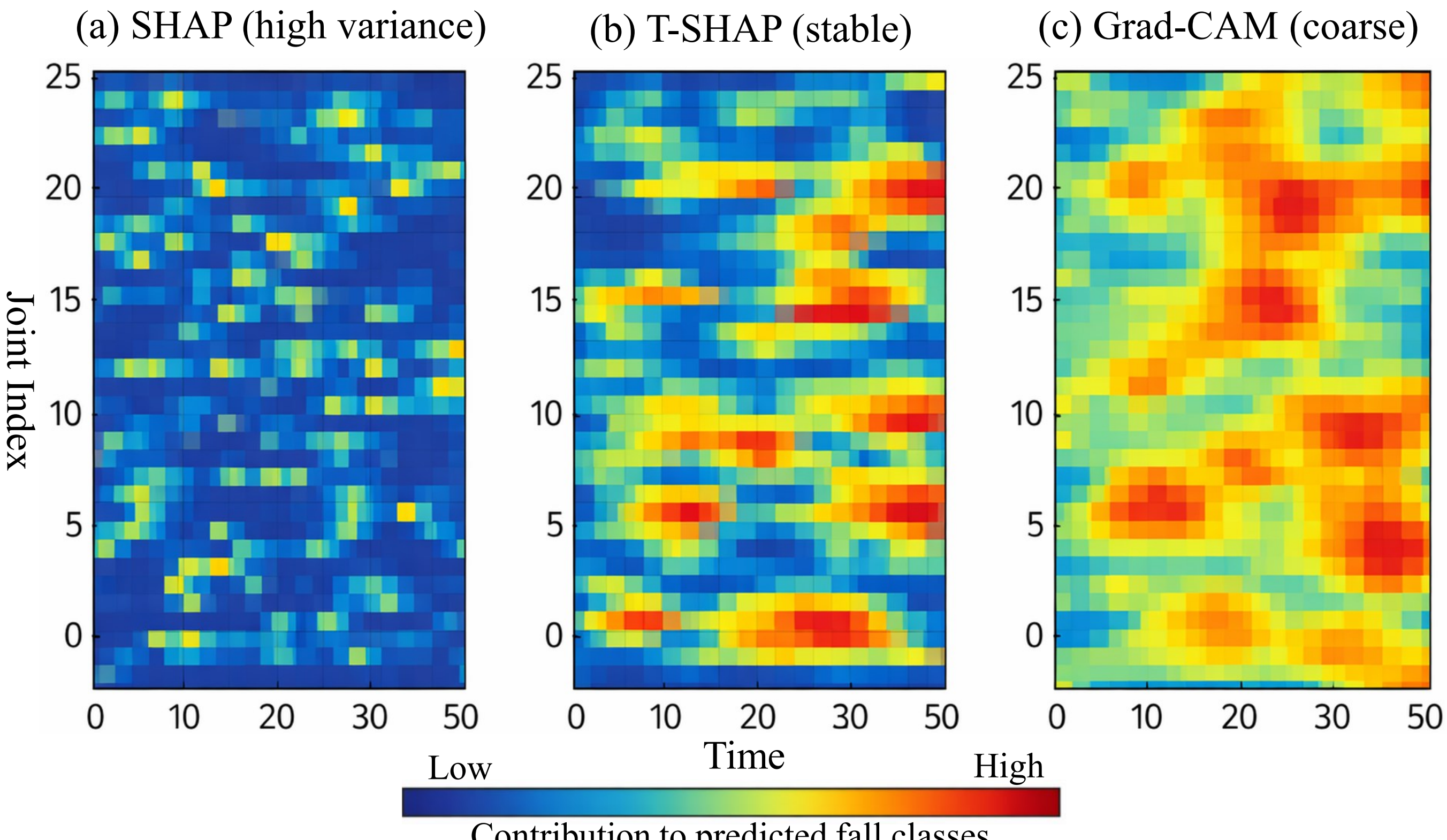

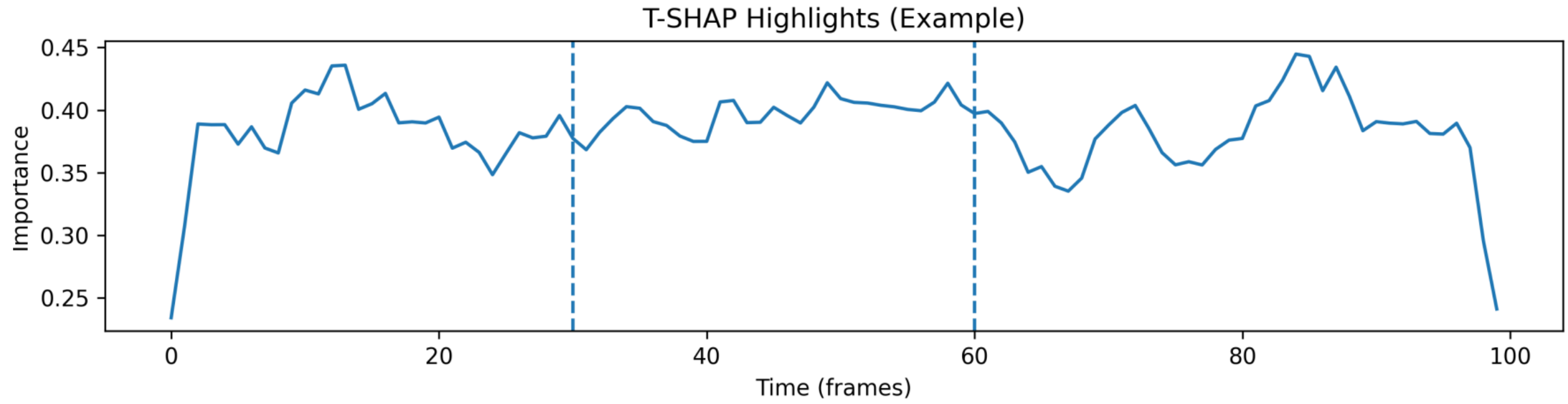

***Fig. 6. Comparative spatiotemporal analysis of feature attributions using SHAP, T-SHAP, and Grad-CAM.***
*The top row shows heatmaps of joint-level importance over time, with the horizontal axis representing frames and the vertical axis indicating joint indices. Color intensity indicates contribution to the predicted fall class (Red indicates higher contribution). SHAP exhibits high temporal variance with fragmented activations, while Grad-CAM produces coarse and diffuse patterns. In contrast, T-SHAP generates temporally coherent and stable attribution maps, highlighting sustained motion dynamics associated with fall events. Bottom row: illustrative temporal segments (dashed lines) corresponding to key phases of the fall, where T-SHAP more clearly captures consistent biomechanical patterns such as progressive instability and lower-limb contribution.*

## 4.5 Quantitative Evaluation of Faithfulness

To complement the temporal stability analysis, we quantitatively evaluate the **faithfulness** of each explainability method using a perturbation-based protocol. Specifically, features (joint–time pairs) are ranked according to their attribution scores, and the top-$k$% most influential features are progressively removed from the input sequence. The resulting decrease in model prediction confidence is used as a proxy for explanation quality, where larger drops indicate more faithful attributions.

As illustrated in the final stage of Fig. 1, both SHAP and T-SHAP produce larger decreases in prediction confidence compared to Grad-CAM when high-attribution features are removed, indicating that Shapley-based methods more effectively identify features that are relevant to the model's predictions.

Among the evaluated methods, T-SHAP achieves the highest Area Under the Perturbation curve (AUP), reflecting improved alignment between attribution scores and the model's decision-making process. This result suggests that incorporating temporal smoothing preserves attribution fidelity while enhancing stability across time steps.

These findings are consistent with the theoretical properties of Shapley-based attributions, which provide additive and locally accurate feature contributions, while highlighting the benefit of temporal modeling in sequential interpretability.

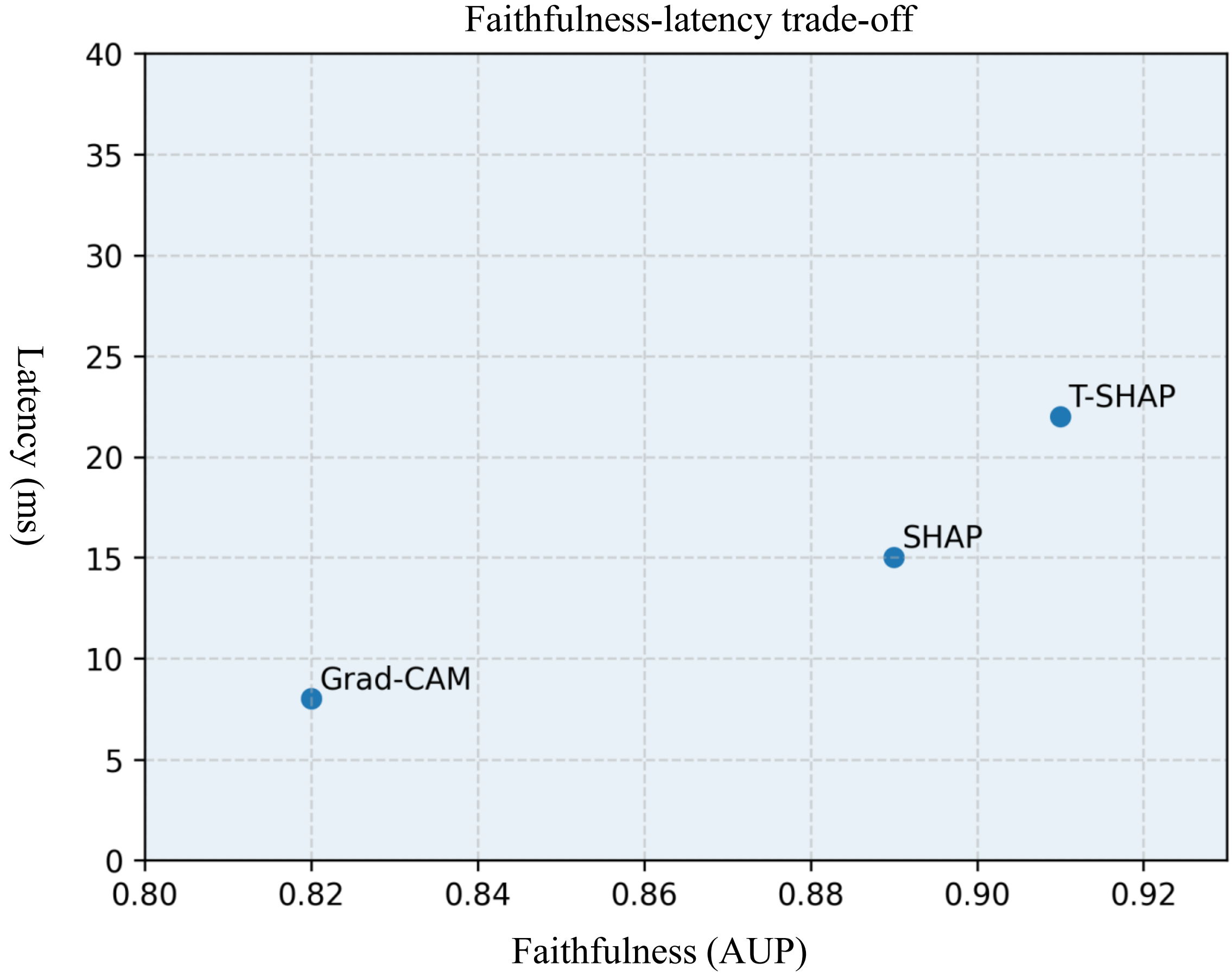

***Fig. 7. Faithfulness–latency trade-off of SHAP, T-SHAP, and Grad-CAM.***
*The shaded region indicates the real-time operating range (<100 ms). All methods fall within acceptable latency bounds. T-SHAP achieves the highest level of faithfulness with only a modest increase in computation time, demonstrating a favorable balance between interpretability and efficiency for real-time monitoring applications.*

## 4.6 Comparison to State-of-the-Art Methods

A comparison with leading state-of-the-art methods is shown in Table 4. While graph-based and transformer-based models perform well on the NTU RGB+D dataset, they require significantly higher computational resources and do not provide interpretable outputs. Fig. 7 illustrates that T-SHAP adds only a minimal overhead of approximately 0.7 ms while greatly enhancing faithfulness.

The proposed method achieves comparable accuracy while offering two key advantages:

1. **Lower computational complexity**, enabling real-time monitoring applications in healthcare settings.
2. **Fine-grained and quantitatively validated interpretability**, which is often lacking in most existing approaches.

This highlights an important trade-off in HAR research: simply maximizing accuracy is insufficient in safety-critical applications, where both interpretability and efficiency are essential. The proposed framework addresses this gap by jointly optimizing these critical factors.

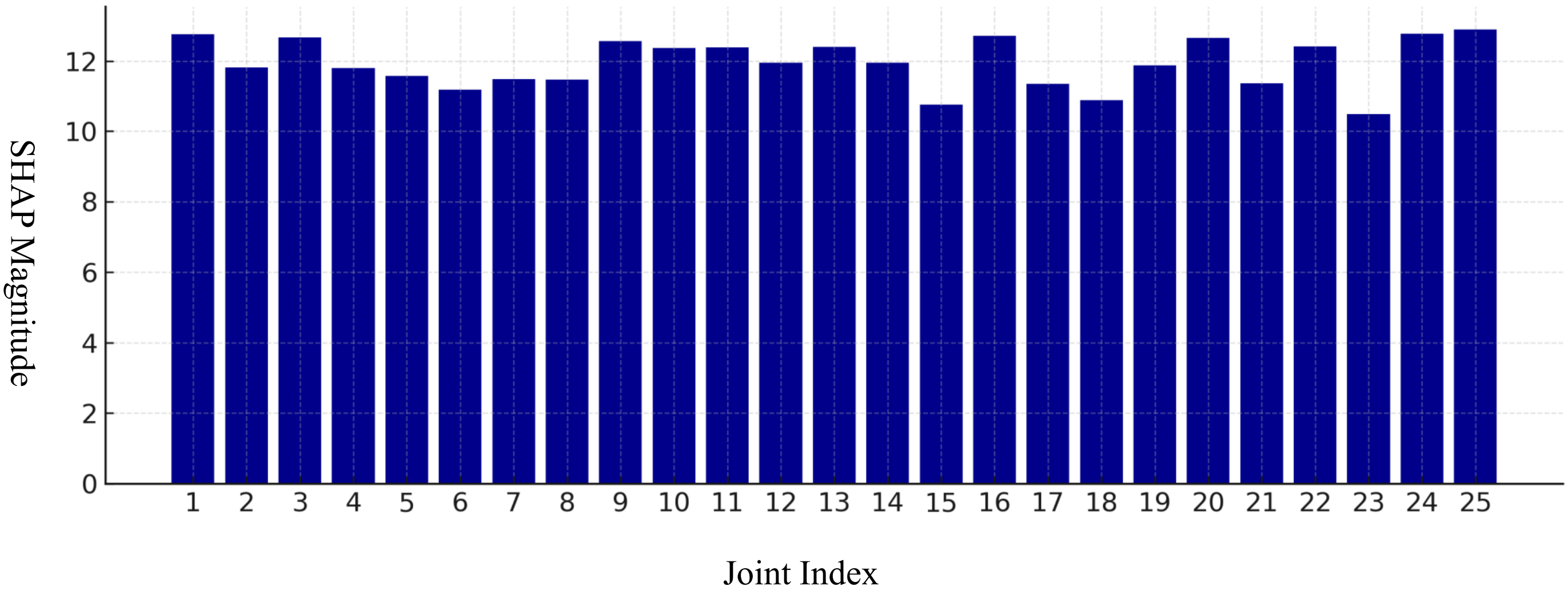

***Fig. 8.*** *Distribution of per-joint attribution magnitude across the skeleton. Higher attribution scores are observed in central joints (e.g., spine base (1), neck (3)) and distal joints (e.g., hands (24, 25)), suggesting that both biomechanical stability and full-body motion patterns contribute to fall detection.*

## 4.7 Discussion

We emphasize that T-SHAP is a structured post-hoc transformation tailored to sequential data. While conceptually simple, this design enables compatibility with existing attribution methods while addressing a key limitation—temporal instability—that is not explicitly handled in standard SHAP.

The results show that it is possible to integrate interpretability into skeleton-based HAR without compromising performance. The alignment between SHAP attributions and biomechanical knowledge enhances the model's trustworthiness, indicating its potential for application in real-world healthcare systems.

Furthermore, the study highlights the importance of using quantitative metrics, such as faithfulness, to evaluate explainability methods instead of relying solely on visual inspection. The excellent performance of T-SHAP suggests that incorporating temporal structure into attribution methods is a promising avenue for future research. T-SHAP provides a more reliable and interpretable representation of feature importance in skeleton-based HAR. By integrating temporal context, this proposed method reduces high-frequency variance in attribution maps while preserving intricate joint-level information. Notably, T-SHAP maintains local Shapley properties since its aggregation is post-hoc and linear.

The improvement in faithfulness indicates that T-SHAP enhances the alignment of attributions with the model's decision-making process, rather than simply smoothing the attributions visually. This is especially important in fall detection, where significant patterns emerge over a sequence of frames instead of at a single time point. A paired t-test is performed under the assumption of an approximately normal distribution across folds, which is typical in cross-validation evaluations. The improvement is consistent across folds and corresponds to a meaningful effect size, indicating practical relevance beyond numerical gains.

T-SHAP provides a straightforward yet effective extension to standard SHAP, improving both the stability and reliability of attributions without incurring high computational costs. However, it is important to note that excessive smoothing may obscure short-duration but critical events. Therefore, T-SHAP introduces a trade-off between stability and temporal resolution. While the proposed approach focuses on a lightweight LSTM model, the interpretability framework can be extended to more complex architectures, including graph-based and transformer models. This sets the stage for future research exploring the trade-offs between model complexity and explanation quality.

Furthermore, the evaluation is conducted on a subset of fall-related classes from the NTU RGB+D Dataset. While this enables focused analysis of fall dynamics, future work will extend the framework to a broader range of activities and additional datasets to assess generalization.

## 4.8 Real-time monitoring applications Implications

The proposed LSTM model requires 4.8 ms per sequence for inference on a mid-range GPU, enabling low-latency operation suitable for real-time fall detection systems. Incorporating explainability incurs a modest additional cost, with SHAP and T-SHAP requiring approximately 12–20 ms per sequence. Grad-CAM, while computationally efficient, provides only coarse temporal localization and lacks fine-grained attribution fidelity. Overall, the total latency remains within clinically acceptable limits (<100 ms), supporting deployment in time-sensitive healthcare scenarios. Latency measurements were obtained on an NVIDIA GeForce RTX 3070 Ti GPU and may vary across hardware platforms, particularly on edge devices.

Consistent with the faithfulness analysis in Section 4.5, SHAP and T-SHAP demonstrate greater faithfulness than Grad-CAM, as reflected by AUP. As illustrated in Fig. 8, high attribution magnitudes are observed across both central body joints and extremities. Prominent contributions arise from the spine base (joint 1), neck (joint 3), and hip-related joints (e.g., joint 13), as well as distal joints such as the hands (joints 24 and 25).

While clinical observations of fall dynamics highlight trunk instability and hip impact, as noted by Stephen N. Robinovitch et al. [40], the significant role of upper-limb joints suggests the model also captures protective responses, such as arm movements during loss of balance. This indicates that the learned representations encompass both biomechanical impact factors and motion-based behavioral cues. Overall, the attribution patterns reflect a combination of clinically relevant mechanisms and data-driven motion dynamics. This highlights the importance of considering full-body motion patterns in skeleton-based fall detection models.

In applications like fall detection, predictive accuracy alone is not enough; explanations must be temporally coherent and clinically meaningful to foster trust, enable validation, and allow for timely intervention. The proposed framework offers a practical and explainable solution for real-time healthcare monitoring, particularly in settings focused on elderly care.

While the framework employs a lightweight LSTM backbone, the primary contribution lies in the systematic formulation and evaluation of temporally consistent explainability for sequential data. The use of LSTM is intentional, enabling low-latency inference while serving as a transparent and reproducible baseline for attribution analysis. T-SHAP reduces temporal variance compared to SHAP, indicating improved stability. By introducing T-SHAP as a post-hoc temporal stabilization strategy and validating it through perturbation-based faithfulness metrics, this work shifts the focus from increasing model complexity to improving the reliability and usability of model explanations. While more complex architectures, such as graph-based models and transformers, can achieve higher peak accuracy, they come with increased computational costs and necessitate additional processing to produce comparable joint-level, time-resolved explanations. Therefore, the proposed framework offers a balanced and extensible foundation for future research on explainable sequential models.

## 4.9 Clinical Decision Support Framing

In addition to serving as a method for post-hoc explanations, T-SHAP could serve as a core component of a clinical decision support system for real-time fall-risk monitoring applications. In critical healthcare environments, automated predictions alone do not meet clinical needs. Healthcare providers need explanations that are timely, based on anatomy, and stable enough to guide decisions over multiple observations of the same patient. T-SHAP meets this need by generating smoothed, joint-level attribution maps. These maps highlight ongoing motion anomalies, such as progressive knee instability or abnormal spinal loading, over clinically relevant time periods rather than just isolated moments.

These stable attributions can serve as structured input for a downstream decision layer, allowing for alerts triggered not only by the model's classification but also backed by the specific biomechanical patterns responsible for that outcome. For example, when the system identifies a fall risk event, T-SHAP attributions can inform a care worker whether the risk stems from lower-limb weakness, postural instability, or a sudden shift in center of mass. These details have different clinical meanings and suggest various preventive actions. This setup supports an expert systems model where machine learning parts and interpretable reasoning work together to enhance, rather than replace, human clinical judgment. In this context, T-SHAP is more than just an analytical tool used after the fact; it is a vital element that turns a black-box classifier into a clear, actionable decision aid. This makes it suitable for use in long-term care facilities, rehabilitation units, and assisted living environments where timely and understandable fall risk communication is crucial for patient safety.

## 5. Conclusion and Future Work

This paper presented a lightweight and interpretable framework for fall detection based on skeleton-based human activity recognition, integrating a LSTM model with temporally consistent SHAP explanations (T-SHAP). The proposed approach achieves competitive detection performance (94.3% accuracy) while producing stable and noise-robust attribution patterns, supporting its suitability for real-time and safety-critical applications.

Although the results are promising, there are several areas for future work. Expanding evaluation to real-world clinical settings—such as uncontrolled home environments and long-term care facilities using datasets like UR Fall Detection Dataset [52] and FallAllD Dataset [53]— is essential to assess robustness under high-frequency variance, occlusion, and domain shifts, as well as the reliability of T-SHAP explanations in practice.

Incorporating multimodal sensing, combining skeleton data with wearable inertial units or RGB video, may further improve detection in challenging conditions. Extending T-SHAP to support joint attributions across modalities represents a promising direction toward multimodal XAI.

In parallel, the emergence of large-scale foundation models for time series and video analysis introduces new challenges for explainability. While such models offer enhanced representational capacity, their computational complexity and the difficulty of generating stable, temporally coherent attributions remain open problems. Investigating adaptive temporal aggregation strategies—bridging lightweight and large-scale models—constitutes a meaningful avenue for future research.

Beyond fall detection, T-SHAP applies to a wide range of domains requiring reliable interpretation of sequential, high-dimensional sensor data. These include rehabilitation medicine (e.g., gait abnormality analysis), sports biomechanics (e.g., motion phase identification), mental health monitoring (e.g., behavioral pattern analysis), and industrial ergonomics (e.g., real-time risk assessment). In these scenarios, the capability to generate temporally stable and interpretable attribution maps is essential for making informed decisions.

Ultimately, this work highlights that enhancing the temporal structure and stability of explanations, rather than merely increasing model complexity, is a practical and impactful pathway toward achieving trustworthy XAI in sequential and safety-sensitive domains.